\documentclass[10pt, a4paper]{article}
\usepackage{lrec}
\usepackage{multibib}
\newcites{languageresource}{Language Resources}
\usepackage{graphicx}
\usepackage{tabularx}
\usepackage{soul}

\usepackage{rotating}

\usepackage{epstopdf}
\usepackage[utf8]{inputenc}

\usepackage{hyperref}
\usepackage{xstring}
\usepackage{footnote}
\usepackage{enumerate}

\usepackage{CJKutf8}
\newenvironment{SChinese}{%
  \CJKfamily{gbsn}%
  \CJKtilde
  \CJKnospace}{}

\newcolumntype{C}{>{\centering\arraybackslash}X}
\newcolumntype{L}{>{\raggedright\arraybackslash}X}

\title{Evaluating Machine Translation Performance on Chinese Idioms\\ with a Blacklist Method}

\name{Yutong Shao$^{1}$, Rico Sennrich$^{2}$, Bonnie Webber$^{2}$, Federico Fancellu$^{2}$}

\address{$^{1}$School of Electronic Engineering and Computer Science, Peking University \\ 5 Yiheyuan Road, Haidian District, Beijing, China \\ sythello@pku.edu.cn \\ \\
	$^{2}$School of Informatics, University of Edinburgh \\ 10 Crichton Street, Edinburgh, United Kingdom, \\ rico.sennrich@ed.ac.uk, bonnie@inf.ed.ac.uk, s1260346@sms.ed.ac.uk}

\abstract{
Idiom translation is a challenging problem in machine translation because the meaning of idioms is non-compositional, and a literal (word-by-word) translation is likely to be wrong. In this paper, we focus on evaluating the quality of idiom translation of MT systems. We introduce a new evaluation method based on an idiom-specific \emph{blacklist} of literal translations, based on the insight that the occurrence of any blacklisted words in the translation output indicates a likely translation error. We introduce a dataset, CIBB (Chinese Idioms Blacklists Bank), and perform an evaluation of a state-of-the-art Chinese$\to$English neural MT system. Our evaluation confirms that a sizable number of idioms in our test set are mistranslated (46.1\%), that literal translation error is a common error type, and that our blacklist method is effective at identifying literal translation errors.\newline
\Keywords{Chinese-English machine translation, evaluation, idiom translation, blacklist method, CIBB dataset}
}

\begin{document}

\maketitleabstract

\section{Introduction}
\label{Introduction}
Idioms are a special figure of speech that are non-compositional and non-literal, though occasionally share surface realizations with literal language uses \cite{DBLP:conf/mwe/SaltonRK14}. Idioms are considered highly problematic for a wide variety of NLP tasks \cite{DBLP:conf/cicling/SagBBCF02}. This belief also holds true for machine translation, because MT systems often make the assumption that meaning is compositional, which is not true for idioms. The compositionality assumption leads to \textit{literal translation errors}, the word-by-word translation of idioms, resulting in a translation that is confusing and not understandable. Therefore, idiom translation is a hard problem in MT and has attracted considerable research interest \cite{DBLP:conf/mwe/CapNWW15,DBLP:conf/mwe/SaltonRK14,DBLP:books/daglib/0029210}.

Given the difficulty of idiom translation in MT, it would be helpful to have a method to evaluate idiom translation performance.
There is a wide range of methods for evaluating the performance of MT systems, but none of them are satisfactory for the targeted evaluation of idiom translation.
The most straightforward method is human evaluation.
While human evaluation is highly valuable, it is desirable to develop complementary automatic methods that are low-cost and fast, thus allowing for more rapid and frequent feedback cycles.
Popular automatic MT metrics such as BLEU \cite{DBLP:conf/acl/PapineniRWZ02} are inexpensive, but are unsuitable for a targeted evaluation.

This paper tries to fill this gap by presenting a method to assess the quality of idiom translations. We introduce a new method called ``blacklist method'' for performance evaluation on idioms, which is based on the intuition that a literal translation of the components of the idiom is likely to be wrong, and easy to spot by defining a blacklist of words that indicate a likely literal translation error. 

We perform a case study on a special class of Chinese idioms that typically consist of 4 characters, called ``cheng2 yu3''.
Actually, not all these 4-character words satisfy the definition of idioms. Some words are \textit{semantically transparent}, which means they are compositional and can be translated literally. This kind of words are less problematic and less necessary to evaluate the systems' performance on. In this research we will only focus on those \textit{semantically non-transparent} words, which have different literal meanings and idiomatic meanings. We will subsequently refer to them as ``Chinese idioms''.\newline
We also introduce the CIBB dataset\footnote{This dataset is released at \texttt{https://github.com/sythello/CIBB-dataset}} for actually executing this evaluation on Chinese$\to$English MT systems. Based on this dataset, we conduct experiments on a state-of-the-art NMT system. From the experiments we draw the following conclusions:
\begin{enumerate}
\item Idiom translation remains an open problem in Chinese$\to$English NMT
\item Literal translation error is still a prevalent error type
\item The blacklist method is effective at detecting literal translation errors.
\end{enumerate}

\section{Related Work}

\subsection{Global Evaluation Metrics}
Global evaluation metrics are metrics that evaluate the overall performance of MT systems and allow automatically calculation. There are many well-known global evaluation metrics, such as BLEU \cite{DBLP:conf/acl/PapineniRWZ02}, METEOR \cite{banerjee2005meteor}, TER \cite{snover2006study}, etc. However, these metrics only provide global evaluation and are unable to evaluate MT systems' performance on specific aspects. Therefore, they are unsatisfactory in evaluating idiom translation performance.

\subsection{Test Suite Methods}
\label{Test_Suites}
Test suite methods construct a set of sentences that focus on specific types of difficulties in MT. Typically, we design a set of sentences in the source language for the MT system to translate, and a scoring method to evaluate the translations. The sentence set and the scoring method are designed so that the score assigned to a system indicates the system's performance on the focused difficulty. This kind of methods makes up for the drawbacks of global evaluation metrics that they cannot assess a system's performance on specific issues. \newline
There are many previous works belonging to this category. \cite{D17-1262} proposed a \textit{challenge set approach} to evaluate English$\to$French MT systems' performance on divergence problems. The English sentences in the challenge set are chosen so that their closest French equivalent will be structurally divergent from them in some crucial way. \cite{burchardt2017linguistic} constructed a test suite for English$\to$German MT systems. This test suite covers a wide variety of linguistic phenomena, such as ambiguity, composition, function words, multi-word expressions and so on. \cite{DBLP:conf/wmt/BurlotY17} introduced a new scheme to evaluate the performance of English$\to$MRL (morphologically rich languages) MT systems on morphological difficulties. The test suite they built consists of three parts, focusing on a system's morphological adequacy (generating different morphological features in different contexts), fluency (word agreement) and certainty (generating the same morphological features in different contexts), respectively. Evaluation is based on automatic morphological analysis of the MT output.
\cite{DBLP:conf/eacl/Sennrich17} proposed a method to construct the test suite automatically for evaluating English$\to$German NMT systems on word agreement, polarity, transliteration, etc. The test suite is made up with minimal translation pairs,  where a reference translation is paired with a contrastive translation which introduces a single translation error, allowing to measure the sensitivity of a neural MT (NMT) system towards this type of error. The score on the test suite is also obtained automatically by calculating the precision of the NMT system to assign a higher probability to the correct translation than to the contrastive translation in each translation pair.
While this method allows for an automatic large-scale evaluation of specific errors, it only measures the probability of pre-defined translations, and is less suitable if the types of errors are relatively unpredictable. Even if we only focus on literal translation errors, it is hard to align the idiom with its translation in the reference (because the translation of idioms can be very flexible) and replace it with literal translation without destroying the coherence of the whole sentence.

Test suite methods can be divided into manual-construction methods and automatic-construction methods, based on whether the test suite construction is automatic. They can also be divided into manual-evaluation methods and automatic-evaluation methods, based on whether the scoring process is automatic. Combining the two classification criteria, we have 3 typical categories of test suite methods:\newline
\begin{enumerate}
\item Automatic construction, automatic evaluation: large test suites and efficient evaluation.
\item Manual construction, automatic evaluation: small test suites but efficient evaluation.
\item Manual construction, manual evaluation: small test suites and laborious evaluation.
\end{enumerate}
According to the classification criteria given above, \cite{D17-1262} and \cite{burchardt2017linguistic} are manual construction, manual-evaluation test suite methods; \cite{DBLP:conf/wmt/BurlotY17} and \cite{DBLP:conf/eacl/Sennrich17} are automatic-construction, automatic-evaluation test suite methods.\newline

\subsection{Automatic Error Detection Methods}
Automatic error detection methods complement global evaluation metrics in another way, by providing algorithms to detect specific kinds of errors in the translation automatically. Previously, there have been many valuable works on automatic error detection. \cite{Addicter} introduced Addicter, which can detect many translation error types, such as missing word, untranslated word, extra word, form error, etc. It is based on the word alignment between the reference and the hypothesis. \cite{Hjerson} introduced Hjerson, which detects similar error types as Addicter, while it is based on the dynamic programming algorithm for calculating Word Error Rate (WER). However, both Addicter and Hjerson have some drawbacks in common. First, they work on a word-by-word basis, so the error types they can detect are rather restricted. Also, they do not match well enough with human annotators, implied by the experiment results in \cite{Addicter}.

\subsection{Idiom Translation and Literal Translation}
\label{Idiom Translation and Literal Translation}
Idioms have long been considered as a hard problem for machine translation in many language pairs. Experiments in \cite{DBLP:conf/hytra/SaltonRK14} showed that on sentences containing idioms, a standard phrase-based English$\to$Brazilian-Portuguese MT system achieves about half the BLEU score of the same system when applied to sentences that do not contain idioms. 
Among all the translation errors caused by idioms, \textit{literal translation errors} are believed to be an important error type. \cite{DBLP:conf/mipro/ManojlovicDB17} demonstrated that literal translations predominate in the output of a phrase-based English$\leftrightarrow$Croatian MT system when translating sentences with idioms. According to our preliminary observations, literal translation errors also occur often in state-of-the-art Chinese$\to$English NMT systems.\newline 
In order to improve the performance of idiom translation, \cite{carpuat-diab:2010:NAACLHLT} investigate two strategies: treating idioms and multiword expressions as an atomic unit, and adding a phrase-level feature that identifies multiword expressions. They find that both strategies improve the translation of non-compositional expressions.
\cite{DBLP:conf/mwe/SaltonRK14} propose a substitution method that replaces idioms in the source sentence with their literal meaning before translation; after translation, the translation of the literal meaning is replaced with a target language idiom, if possible.

\section{Blacklist Method}

The ``blacklist'' method we are going to describe is used for detecting \textit{literal translation errors}, which means the system translates an idiom word-by-word and thus gets a wrong translation, as described in section \ref{Introduction} According to related works and our own observations introduced in section \ref{Idiom Translation and Literal Translation}, we hypothesize that literal translation errors represent a majority of idiom translation errors.
We further hypothesize that we can easily identify literal translation errors by checking the translation for words that represent the meaning of a subsequence of the source idiom, but which should not appear in the true, idiomatic translation.
These words make up the \textit{blacklist} for the idiom, which we manually create.
For the example in Table \ref{table:example_blacklist}, if a machine translation system is fed with a Chinese sentence containing this idiom, and the system outputs a translation containing ``bamboo'' or ``chest'', then we say the translation \textit{trigger the blacklist} and therefore will be judged as a literal translation error.\newline

\begin{CJK}{UTF8}{}\begin{SChinese}
\begin{table}[!h]
\begin{center}
\begin{tabularx}{\columnwidth}{|C|L|}

      \hline
      Idiom & 胸有成竹\\
      \hline
      Idiomatic translation (correct) & Be very ready; have a well-thought-out plan\\
      \hline
      Literal translation (incorrect) & Have a well-formed bamboo in one's chest\\
      \hline
      \textbf{Blacklist} & \textbf{bamboo, chest}\\
      \hline

\end{tabularx}
\caption{Example of blacklist.}
\label{table:example_blacklist}
 \end{center}
\end{table}
\end{SChinese}\end{CJK}

Using the concept of blacklist, here we give the whole process of ``blacklist method'' evaluation:\newline
\begin{enumerate}
\item Build an idiom list with idioms that we can build blacklist for. To be more specific, we choose idioms that contain one or more characters whose direct translations should not exist in translation of the whole idiom.
\label{buildidiomlist}
\item Build a blacklist for each idiom on the list. The blacklist consists of the direct translation of the characters mentioned in the last step.
\label{buildblacklist}
\item Gather source language (Chinese) sentences containing idioms on the list. Note that the method itself does not need reference translations. Nevertheless, if someone is not a speaker of the source language but wishes to get some ideas about the detected literal translation errors, or to check whether the detection is correct, then using translation pairs is more desirable than monolingual sentences.
\label{buildsentenceset}
\item Feed all the sentences to the MT system to get the translations.
\label{feedtosystem}
\item Calculate the percentage of translations triggering the blacklist, which is the evaluation score for the system.
\label{calculatescore}
\end{enumerate}

We draw on an existing idiom list for step \ref{buildidiomlist}, and perform step \ref{buildblacklist} manually. Steps \ref{buildidiomlist}-\ref{buildsentenceset} form the construction procedure and only need to be conducted once; steps \ref{feedtosystem}-\ref{calculatescore} form the evaluation procedure that needs to be conducted on different systems.
\paragraph{Advantages and Disadvantages} According to the classification criteria introduced in section \ref{Test_Suites}, our blacklist method is a manual-construction, automatic-evaluation test suite method. Therefore, the main advantage of the blacklist method is that, after creating the blacklist, large-scale evaluation is inexpensive and reproducible.
The selection of proper idioms and the construction of a blacklist for each idiom is feasible by a bilingual speaker, and future work may even try to automate this.
After the idiom list and blacklists are determined, we can scale up the set of translation pairs as much as we need, using online bilingual or even monolingual datasets. 
Also, we expect the blacklist method to achieve a high precision, because the definition of ``blacklist'' is actually closely related to literal translation errors.
On the other hand, the drawback of this method is that the method is restricted to only one error type, literal translation errors, and will not detect any other type of errors such as deletions or repetitions of the idiom. Hence, recall is uncertain.

\section{Dataset Construction}
In CIBB, we provide a list of 50 Chinese idioms, each paired with an idiom-specific blacklist, and 1194 Chinese$\to$English translation pairs, each containing an idiom on the list.

\paragraph{Idioms and Blacklists}
We downloaded about 30000 Chinese idioms from the following websites:
\begin{itemize}
\item http://www.gsdaquan.com
\item http://chengyu.t086.com
\item http://bcc.blcu.edu.cn
\end{itemize}
After excluding all the idioms that never appeared in the training data of our NMT system, there are about 9000 idioms left. Among these 9000 idioms, we observed some samples of them and selected 50 idioms with different frequencies in the training data. According to our observation, idioms with very high frequency in the training data are generally translated well, so we focus on lower-frequency idioms. Meanwhile, we cannot expect a system to learn to translate idioms with too low frequency. Therefore, we selected idioms appearing between 7 and 1000 times in the training data. We further select only idioms whose translation is non-compositional, and create a blacklist for each idiom.

\paragraph{Translation Pairs}
The translation pairs were extracted from OpenSubtitles2016 dataset \cite{DBLP:conf/lrec/LisonT16}, where we searched for Chinese$\to$English translation pairs with idioms on our list. In order to balance the frequency of all the idioms in the translation pairs, preventing the majority being taken up by only a few idioms, we restricted the maximum occurrences of any idiom to be 40. Under such restrictions, we extracted a total of 1194 translation pairs.

\section{Experiments}

The objective of our experiments is to evaluate the effectiveness of the blacklist method at detecting translation errors, especially the literal translation errors, by its precision, recall, as well as the correlation with BLEU. Also, we want to test to what extent idiom translation is a problem for a current state-of-the-art NMT system.

\subsection{The MT System}
\label{Word_Segmentation}
As a representative of the current state of the art in NMT, we evaluate the Edinburgh NMT system for the WMT17 shared news translation task \cite{DBLP:conf/wmt/SennrichBCGHHBW17}, which was ranked tied best for Chinese$\to$English.
The system is an attentional encoder-decoder, and its training data is constrained to the training data provided at WMT17, namely \emph{News Commentary v12}, \emph{UN Parallel Corpus V1.0}, the \emph{CWMT Corpus}, and back-translated monolingual data from the \emph{News Crawl Corpus}.
On the Chinese side, the system uses Jieba\footnote{\url{https://github.com/fxsjy/jieba}} for word segmentation, and BPE for subword segmentation \cite{sennrich-haddow-birch:2016:P16-12}.
More details about the model architecture can be found in the system description.

\paragraph{Word Segmentation}
Our MT system performs both word segmentation and subword segmentation on the Chinese texts. 
It is worth noting that different approaches of word segmentation may lead to different results in our test.
The test method focuses on literal translation errors, which can only happen if an idiom is segmented into several parts, not if the idiom is unsegmented and treated as a single unit.
Treating the idiom as a single unit may be an effective approach to prevent literal translation errors, but may increase vocabulary size and/or cause other types of errors that are not captured by the blacklist.
Evaluating the effect of (sub)word segmentation on idiom translation remains the subject of future work.

\subsection{Experiment Setup}
We first translate all the 1194 Chinese source sentences into English using the Edinburgh WMT17 system introduced above. Then we apply blacklist method to all the translations. For those translations triggering the blacklist, we manually count the number of correct and incorrect translations, as well as the number of literal translation errors. For those not triggering the blacklist, we randomly sample and manually evaluate 100 translations to estimate error rates for this group. We only focus on errors with respect to idioms; errors of other aspects are ignored.

\subsection{Experiment Results}
\label{Test_Results}
Among all the 1194 translations, 145 triggered the blacklist and 1049 translations did not. We conducted manual evaluation on all the translation triggering the blacklist, and 100 random sampled translations not triggering the blacklist. The results are shown in Table \ref{table:test_results}.

\begin{table}[!h]
\begin{center}
\begin{tabularx}{\columnwidth}{|c|C|C|C|C|}

      \hline
       & Correct & Incorrect & Incorrect Literal & Total\\
      \hline
      Not triggering & 640* & 409* & 0* & 1049\\
      \hline
      Triggering & 3 & 142 & 142 & 145\\
      \hline
      Total & 643 & 551 & 142 & 1194\\
      \hline
      
\end{tabularx}
\caption{Results of our test on Edinburgh WMT17 system. Figures followed by (*) are estimated using 100 random samples out of 1049.}
\label{table:test_results}
 \end{center}
\end{table}

First of all, the overall 46.1\% (551/1194) error rate and 11.9\% (142/1194) literal translation error rate implies that idiom translation is still problematic for a state-of-the-art MT system and literal translation is an important error type in idiom translation. Furthermore, among the 145 translations triggering the blacklist, 142 were incorrect literal translations; only 3 of them were actually correct ones but triggered the blacklist in some other ways (an example of this is provided in section \ref{False Positives}). For the translations not triggering the blacklist, according to our evaluation on 100 examples, 61 of them were correct and 39 were incorrect, while no literal translation errors was found. We thus estimated that for all the 1049 translations not-triggering the blacklist, 640 are correct translations while 409 are incorrect, and there is no literal translation error. This means that our blacklist method has a very high precision of 97.9\% (142/145) and recall of 100\% (142/142) of catching literal translation errors. If we regard the blacklist method as a method to detect wrong idiom translations of any type, the precision is unchanged, and we still have a recall of about 25.8\% (142/551), which means the blacklists can catch a considerable amount of errors in all the translations.
Among errors that the blacklist method does not identify, deletion errors are the most prevalent category.

\subsection{Idiom Translation and BLEU}

We test the interaction of our evaluation method and BLEU
We calculated the BLEU score for four different sets of translations: A random sample of 1000 sentences from OpenSubtitles2016, our CIBB test set of 1194 sentences containing an idiom, all translations triggering the blacklist and all translations not triggering the blacklist. The results are listed in Table \ref{table:BLEU_scores}. We can see that the BLEU score for translations of idioms is only about half the BLEU score of randomly sampled translations, in line with results from previous work \cite{DBLP:conf/hytra/SaltonRK14}. This confirms our hypothesis that translating sentences with idioms is hard for state-of-the-art NMT systems. Also, the BLEU score of translations triggering the blacklist is lower than the translations not triggering the blacklist, indicating that the blacklist method is useful at identifying low-quality translations, even without a reference.

\begin{savenotes}
\begin{table}[!h]
\begin{center}
\begin{tabular}{|l|r|}

      \hline
      test set & BLEU\\
      \hline
      Random 1000 samples & 11.85\\
      \hline
      With idioms & 6.35\\
      \hline
      Blacklist triggered & 5.64\\
      \hline
      Blacklist not triggered & 6.44\\
      \hline

\end{tabular}
\caption{BLEU scores for different sets of translations.}
\label{table:BLEU_scores}
 \end{center}
\end{table}
\end{savenotes}

\subsection{Examples}
Here we provide some examples for different types of translations we discussed in section 4.2.
\subsubsection{Correctly Detected Errors}

\begin{CJK}{UTF8}{}
\begin{SChinese}

\begin{table}[!h]
\begin{center}
\begin{tabularx}{\columnwidth}{|C|L|}

\hline
Idiom & 说三道四\\
\hline
Meaning & Gossip\\
\hline
Literal & Speak three and four\\
\hline
Blacklist & \textbf{three} four\\
\hline
SRC & 医生说了你不能对我说三道四\\
\hline
REF & The therapist said you're not allowed to judge me.\\
\hline
TRANS & The doctor said that you can't say \textbf{three} things to me.\\
\hline

\end{tabularx}
\caption{Example for correctly detected errors.}
\label{table:example1}
 \end{center}
\end{table}


In the example shown in Table \ref{table:example1}, the word ``three'' is the literal translation of 三, but should not appear in the correct idiomatic translation
Therefore, the occurrence of ``three'' in the translation triggers our blacklist, correctly indicating a literal translation error.

\subsubsection{False Positives}
\label{False Positives}

\begin{table}[!h]
\begin{center}
\begin{tabularx}{\columnwidth}{|C|L|}

\hline
Idiom & 谈笑风生\\
\hline
Meaning & Talk cheerfully and humorously\\
\hline
Literal & Talking and laughing generate winds\\
\hline
Blacklist & \textbf{wind}\\
\hline
SRC & 他们谈笑风生 而我们却要在这里吹风\\
\hline
REF & Burke's up there, too laughing it up with the President while we're stuck down here.\\
\hline
TRANS & They \textit{talk and laugh}, but we're going to blow the \textit{\textbf{wind}} right here\\
\hline

\end{tabularx}
\caption{Example for false positives.}
\label{table:example2}
 \end{center}
\end{table}

The example shown in Table \ref{table:example2} demonstrates a false positive. While the idiom is translated correctly into ``talk and laugh'', ``wind'' appears in another place of the source sentence, and that triggered the blacklist.
Future work could involve further constraints, such as taking into account alignment information, to further reduce false positives.

\subsubsection{Not Detected Errors}

\begin{table}[!h]
\begin{center}
\begin{tabularx}{\columnwidth}{|C|L|}

\hline
Idiom & 生龙活虎\\
\hline
Meaning & Full of energy\\
\hline
Literal & Lively dragon and tiger \\
\hline
Blacklist & dragon tiger\\
\hline
SRC & 你明明生龙活虎到处走\\
\hline
REF & You were so actively walking around just then\\
\hline
TRANS & You \textbf{have to} go all over the place\\
\hline

\end{tabularx}
\caption{Example for not detected errors.}
\label{table:example3}
 \end{center}
\end{table}

In this example shown in Table \ref{table:example3}, the idiom meaning ``full of energy'' or ``actively'' is incorrectly translated into ``have to''. However, as this is not a literal translation error, our blacklist method is unable to catch it.
This is a limitation of the blacklist method, which is only designed to capture literal translation errors.

\end{SChinese}
\end{CJK}

\section{Conclusion and Future Work}

We introduced the blacklist method for evaluating the performance of MT systems on idioms. This method works by automatically detecting literal translation errors and calculating the error rate. The results of our experiments have shown that the blacklist method is useful for detecting this kind of errors. The experiments also confirm that idiom translation remains an open problem for NMT systems. We introduced the dataset CIBB which is used for executing blacklist method evaluation on Chinese$\to$English MT systems. The dataset contains 1194 Chinese$\to$English translation pairs covering 50 Chinese idioms.\newline
In the future, this work may be developed in following directions:\newline 
\begin{itemize}
\item Our current idiom list consists of 50 idioms, and we can further extend the idiom list and refine the blacklist to improve the performance of the blacklist evaluation method.
\item An automatic identification of idioms, and automatic construction of the blacklist would facilitate the transfer of the evaluation method to other language pairs. We note that there is related work on automatic identification of non-compositional expressions that could enable this \cite{DBLP:conf/emnlp/Melamed97}.
\item While a blacklist-based evaluation has shown high precision and recall at identifying literal translation errors, it is blind towards other error types, such as deletion errors.
We note that related research has focused on the identification and prevention of deletion errors via measuring the ability of models to reconstruct the source sentence from the translation \cite{DBLP:journals/corr/LiJ16,DBLP:conf/aaai/TuLSLL17}.
We consider it interesting that reconstruction-based methods may be blind towards literal translation errors, which means that these two methods are complementary and could potentially be combined.
\end{itemize}

More broadly, a blacklist-based evaluation is attractive in that it can identify some types of translation errors without access to human reference translation. It could thus prove beneficial for quality estimation in a post-editing environment.
Finally, we hope that our evaluation results and dataset will spark future research on improving idiom translation in MT.
We could revisit strategies from phrase-based MT, such as forcing idioms to be represented as an atomic unit \cite{carpuat-diab:2010:NAACLHLT},
although this would have undesirable side effects in neural MT such as increasing the size of the network vocabulary.

\section{Acknowledgements}

This research is supported by the Summer Research Visitor Programme between the School of Electronic Engineering and Computer Science at Peking University and the Institute for Language, Cognition and Computation in the School of Informatics at the University of Edinburgh.

\section{Bibliographical References}
\label{main:ref}

\bibliographystyle{lrec}
\bibliography{xample}

\begin{thebibliography}{}

\bibitem[\protect\citename{Anastasiou}2010]{DBLP:books/daglib/0029210}
Anastasiou, D.
\newblock (2010).
\newblock {\em {Idiom Treatment Experiments in Machine Translation}}.
\newblock Cambridge Scholars Publishing.

\bibitem[\protect\citename{Banerjee and Lavie}2005]{banerjee2005meteor}
Banerjee, S. and Lavie, A.
\newblock (2005).
\newblock {METEOR: An automatic metric for MT evaluation with improved
  correlation with human judgments}.
\newblock In {\em {Proceedings of the acl workshop on intrinsic and extrinsic
  evaluation measures for machine translation and/or summarization}},
  volume~29, pages 65--72.

\bibitem[\protect\citename{Burchardt \bgroup et al.\egroup
  }2017]{burchardt2017linguistic}
Burchardt, A., Macketanz, V., Dehdari, J., Heigold, G., Peter, J.-T., and
  Williams, P.
\newblock (2017).
\newblock {A Linguistic Evaluation of Rule-Based, Phrase-Based, and Neural MT
  Engines}.
\newblock {\em The Prague Bulletin of Mathematical Linguistics},
  108(1):159--170.

\bibitem[\protect\citename{Burlot and Yvon}2017]{DBLP:conf/wmt/BurlotY17}
Burlot, F. and Yvon, F.
\newblock (2017).
\newblock {Evaluating the morphological competence of Machine Translation
  Systems}.
\newblock In {\em {WMT}}, pages 43--55. Association for Computational
  Linguistics.

\bibitem[\protect\citename{Cap \bgroup et al.\egroup
  }2015]{DBLP:conf/mwe/CapNWW15}
Cap, F., Nirmal, M., Weller, M., and {Schulte im Walde}, S.
\newblock (2015).
\newblock {How to Account for Idiomatic German Support Verb Constructions in
  Statistical Machine Translation}.
\newblock In {\em {MWE@NAACL-HLT}}, pages 19--28. The Association for Computer
  Linguistics.

\bibitem[\protect\citename{Carpuat and Diab}2010]{carpuat-diab:2010:NAACLHLT}
Carpuat, M. and Diab, M.
\newblock (2010).
\newblock {Task-based Evaluation of Multiword Expressions: a Pilot Study in
  Statistical Machine Translation}.
\newblock In {\em {Human Language Technologies: The 2010 Annual Conference of
  the North American Chapter of the Association for Computational
  Linguistics}}, pages 242--245, Los Angeles, California, June. Association for
  Computational Linguistics.

\bibitem[\protect\citename{Isabelle \bgroup et al.\egroup }2017]{D17-1262}
Isabelle, P., Cherry, C., and Foster, G.
\newblock (2017).
\newblock {A Challenge Set Approach to Evaluating Machine Translation}.
\newblock In {\em {Proceedings of the 2017 Conference on Empirical Methods in
  Natural Language Processing}}, pages 2476--2486. Association for
  Computational Linguistics.

\bibitem[\protect\citename{Li and Jurafsky}2016]{DBLP:journals/corr/LiJ16}
Li, J. and Jurafsky, D.
\newblock (2016).
\newblock {Mutual Information and Diverse Decoding Improve Neural Machine
  Translation}.
\newblock {\em CoRR}, abs/1601.00372.

\bibitem[\protect\citename{Lison and Tiedemann}2016]{DBLP:conf/lrec/LisonT16}
Lison, P. and Tiedemann, J.
\newblock (2016).
\newblock {OpenSubtitles2016: Extracting Large Parallel Corpora from Movie and
  {TV} Subtitles}.
\newblock In {\em {LREC}}. European Language Resources Association {(ELRA)}.

\bibitem[\protect\citename{Manojlovic \bgroup et al.\egroup
  }2017]{DBLP:conf/mipro/ManojlovicDB17}
Manojlovic, M., Dajak, L., and Bakaric, M.~B.
\newblock (2017).
\newblock {Idioms in state-of-the-art Croatian-English and English-Croatian
  {SMT} systems}.
\newblock In {\em {MIPRO}}, pages 1546--1550. {IEEE}.

\bibitem[\protect\citename{Melamed}1997]{DBLP:conf/emnlp/Melamed97}
Melamed, I.~D.
\newblock (1997).
\newblock {Automatic Discovery of Non-Compositional Compounds in Parallel
  Data}.
\newblock In {\em {EMNLP}}. {ACL}.

\bibitem[\protect\citename{Papineni \bgroup et al.\egroup
  }2002]{DBLP:conf/acl/PapineniRWZ02}
Papineni, K., Roukos, S., Ward, T., and Zhu, W.
\newblock (2002).
\newblock {Bleu: a Method for Automatic Evaluation of Machine Translation}.
\newblock In {\em {ACL}}, pages 311--318. {ACL}.

\bibitem[\protect\citename{Popovic}2011]{Hjerson}
Popovic, M.
\newblock (2011).
\newblock {Hjerson: An Open Source Tool for Automatic Error Classification of
  Machine Translation Output}.
\newblock {\em Prague Bull. Math. Linguistics}, 96:59--68.

\bibitem[\protect\citename{Sag \bgroup et al.\egroup
  }2002]{DBLP:conf/cicling/SagBBCF02}
Sag, I.~A., Baldwin, T., Bond, F., Copestake, A.~A., and Flickinger, D.
\newblock (2002).
\newblock {Multiword Expressions: {A} Pain in the Neck for {NLP}}.
\newblock In {\em {CICLing}}, volume 2276 of {\em {Lecture Notes in Computer
  Science}}, pages 1--15. Springer.

\bibitem[\protect\citename{Salton \bgroup et al.\egroup
  }2014a]{DBLP:conf/hytra/SaltonRK14}
Salton, G., Ross, R.~J., and Kelleher, J.~D.
\newblock (2014a).
\newblock {An Empirical Study of the Impact of Idioms on Phrase Based
  Statistical Machine Translation of English to Brazilian-Portuguese}.
\newblock In {\em {HyTra@EACL}}, pages 36--41. The Association for Computer
  Linguistics.

\bibitem[\protect\citename{Salton \bgroup et al.\egroup
  }2014b]{DBLP:conf/mwe/SaltonRK14}
Salton, G., Ross, R.~J., and Kelleher, J.~D.
\newblock (2014b).
\newblock {Evaluation of a Substitution Method for Idiom Transformation in
  Statistical Machine Translation}.
\newblock In {\em {MWE@EACL}}, pages 38--42. The Association for Computer
  Linguistics.

\bibitem[\protect\citename{Sennrich \bgroup et al.\egroup
  }2016]{sennrich-haddow-birch:2016:P16-12}
Sennrich, R., Haddow, B., and Birch, A.
\newblock (2016).
\newblock {Neural Machine Translation of Rare Words with Subword Units}.
\newblock In {\em {Proceedings of the 54th Annual Meeting of the Association
  for Computational Linguistics (Volume 1: Long Papers)}}, pages 1715--1725,
  Berlin, Germany, August. Association for Computational Linguistics.

\bibitem[\protect\citename{Sennrich \bgroup et al.\egroup
  }2017]{DBLP:conf/wmt/SennrichBCGHHBW17}
Sennrich, R., Birch, A., Currey, A., Germann, U., Haddow, B., Heafield, K.,
  Barone, A. V.~M., and Williams, P.
\newblock (2017).
\newblock {The University of Edinburgh's Neural {MT} Systems for {WMT17}}.
\newblock In {\em {WMT}}, pages 389--399. Association for Computational
  Linguistics.

\bibitem[\protect\citename{Sennrich}2017]{DBLP:conf/eacl/Sennrich17}
Sennrich, R.
\newblock (2017).
\newblock {How Grammatical is Character-level Neural Machine Translation?
  Assessing {MT} Quality with Contrastive Translation Pairs}.
\newblock In {\em {{EACL} {(2)}}}, pages 376--382. Association for
  Computational Linguistics.

\bibitem[\protect\citename{Snover \bgroup et al.\egroup }2006]{snover2006study}
Snover, M., Dorr, B., Schwartz, R., Micciulla, L., and Makhoul, J.
\newblock (2006).
\newblock {A study of translation edit rate with targeted human annotation}.
\newblock In {\em {Proceedings of association for machine translation in the
  Americas}}, volume 200.

\bibitem[\protect\citename{Tu \bgroup et al.\egroup
  }2017]{DBLP:conf/aaai/TuLSLL17}
Tu, Z., Liu, Y., Shang, L., Liu, X., and Li, H.
\newblock (2017).
\newblock {Neural Machine Translation with Reconstruction}.
\newblock In {\em {AAAI}}, pages 3097--3103. {AAAI} Press.

\bibitem[\protect\citename{Zeman \bgroup et al.\egroup }2011]{Addicter}
Zeman, D., Fishel, M., Berka, J., and Bojar, O.
\newblock (2011).
\newblock {Addicter: What Is Wrong with My Translations?}
\newblock {\em Prague Bull. Math. Linguistics}, 96:79--88.

\end{thebibliography}

\section*{Appendix}

\begin{CJK}{UTF8}{}
\begin{SChinese}
\begin{table*}
\centering

\begin{tabular}{c|l|l}
idiom & translation & literal translation \\
\hline
手无寸铁 & Unarmed & Have no iron in one's hand \\
雪上加霜 & Rub salt into the wound; exacerbate & Frost form on the snow \\
背井离乡 & (Be forced to) leave one's home & Leave the well and hometown \\
五花八门 & Of a wide variety & Five flowers and eight gates \\
立竿见影 & Have an immediate effect & Put up a stick and see the shadow \\
烟消云散 & Disappear, vanish & Vanish like smoke and cloud \\
大刀阔斧 & Bold, drastic, macroscopic, not consider much of the details & Big knife and axe \\
不速之客 & Uninvited guest; unwelcome guest & Not invited (speed) guest \\
冷嘲热讽 & Sarcasm; irony & Cold and hot sarcasm \\
迎刃而解 & (Problem) be easily solved & Break on the knife blade \\
蛛丝马迹 & Traces, clues & Spider silk and horce trace \\
亡羊补牢 & Better late than never & Mend the pen after losing some sheep \\
说三道四 & Gossip & Speak three and four \\
锦上添花 & Embellish what is already beautiful & Add flowers to beautiful cloth \\
马马虎虎 & Careless / Just so-so, passable & Like horses and tigers \\
胆战心惊 & Be terror-stricken & Gall trembling and heart frightened \\
易如反掌 & Very easy, a piece of cake & As easy as turning one's hand \\
开门见山 & Come straight to the point/question & Open door and see the mountain \\
胸有成竹 & Have a well-thought-out plan & Have a bamboo in one's chest \\
蠢蠢欲动 & Be restless to do something; be ready to do something & Be restless to move like worms (stupid) \\
洗耳恭听 & Be all ears; listen carefully & Wash one's ears to listen politely \\
五光十色 & Colorful & Five lights and ten colors \\
九霄云外 & Far, far away & Out of nine clouds \\
推心置腹 & Sincerely; heart-to-heart & Push hearts and settle the stomach \\
谈笑风生 & Talk cheerfully and humorously & Talking and laughing generate winds \\
凤毛麟角 & Extremely rare & Pheonix fur and kylin horn \\
灰飞烟灭 & Vanish; be destroyed (like ashes and smoke) & Ash(grey) fly and smoke vanish \\
星罗棋布 & Spread all over the place & Stars spread and men deployed \\
望尘莫及 & Too far behind to catch up & Only see the dust and cannot catch up \\
天马行空 & In a powerful and unconstrained style & Sky horse traveling in the sky \\
呼之欲出 & Vivid / Coming out soon & Call it and it will show up \\
抛砖引玉 & Make some introductory remarks to set the ball rolling & Throw bricks to attract jades \\
添油加醋 & Add highly coloured details; distort, exaggerate & Add oid and vinegar \\
守株待兔 & Wait around aimlessly for a windfall that is unlikely to come & Wait by a tree for rabbits \\
板上钉钉 & Be fixed; be clinched & Nail on the board \\
顺手牵羊 & Walk off with sth.; steal sth. when walking by & Take away a sheep when walking by \\
呆若木鸡 & Be dumb-struck (as a wooden chicken) & Be dumb as a wooden chicken \\
生龙活虎 & Full of energy & Lively dragon and tiger \\
罄竹难书 & (Crimes) be too numerous to record & Cannot list all even if use up a whole bamboo \\
九牛一毛 & A drop in the ocean & One fur for nine oxen \\
闭门造车 & Carry out one's idea without communicating with the outside & Close the door and make a car \\
老态龙钟 & Very old; senile and doddering & Old like a dragon bell \\
行将就木 & Going to die; one foot in grave & Going to be in the wood \\
鼠目寸光 & Shortsighted & Can only get lights from a short distance, like mice \\
蜻蜓点水 & Scratch the surface & Dragonfly skim the water \\
九死一生 & A slim chance of living; extremely dangerous & Nine deathes, one living \\
鱼龙混杂 & Good and bad things mixed together & Fish and dragons mixed together \\
三六九等 & Various grades and ranks & Three, six or nine levels \\
沾花惹草 & Be promiscuous; flirt around & Touch flowers and play with grasses \\
鸡飞狗跳 & Great disorder; turmoil & Chicken fly and dogs jump \\

\end{tabular}
\caption{Idioms in CIBB with idiomatic and literal translation.}
\end{table*}
\end{SChinese}
\end{CJK}

\begin{CJK}{UTF8}{}
\begin{SChinese}
\begin{table*}
\centering

\begin{tabular}{c|l|r|r|l}
& & \multicolumn{2}{c|}{frequency} & blacklist\\
idiom & blacklist & training & CIBB & trigger rate \\
\hline
手无寸铁 & iron & 1000 & 40 & 0 \\
雪上加霜 & snow frost & 871 & 40 & 0 \\
背井离乡 & well & 717 & 36 & 0 \\
五花八门 & five flower eight door gate & 467 & 21 & 0 \\
立竿见影 & stick shadow & 342 & 11 & 0 \\
烟消云散 & cloud & 341 & 40 & 0 \\
大刀阔斧 & knife axe & 239 & 4 & 0 \\
不速之客 & speed & 225 & 40 & 0 \\
冷嘲热讽 & cold hot & 200 & 40 & 0 \\
迎刃而解 & knife blade & 196 & 42 & 0 \\
蛛丝马迹 & spider horse & 191 & 40 & 0 \\
亡羊补牢 & sheep goat & 189 & 32 & 0.062 \\
说三道四 & three four & 168 & 40 & 0.15 \\
锦上添花 & flower & 167 & 29 & 0 \\
马马虎虎 & horse tiger & 155 & 42 & 0.548 \\
胆战心惊 & gut gall & 151 & 21 & 0 \\
易如反掌 & hand & 147 & 40 & 0 \\
开门见山 & door mountain & 144 & 40 & 0.1 \\
胸有成竹 & chest bamboo & 127 & 35 & 0.143 \\
蠢蠢欲动 & stupid & 102 & 40 & 0.1 \\
洗耳恭听 & wash & 101 & 40 & 0 \\
五光十色 & five ten & 95 & 8 & 0.25 \\
九霄云外 & nine & 88 & 15 & 0 \\
推心置腹 & push stomach belly & 86 & 9 & 0 \\
谈笑风生 & wind & 85 & 12 & 0.083 \\
凤毛麟角 & pheonix kylin & 85 & 3 & 0 \\
灰飞烟灭 & grey fly & 83 & 40 & 0.25 \\
星罗棋布 & star chess & 82 & 1 & 0 \\
望尘莫及 & dust & 79 & 17 & 0.235 \\
天马行空 & sky horse & 74 & 26 & 0.154 \\
呼之欲出 & call & 74 & 16 & 0 \\
抛砖引玉 & brick jade gem stone & 71 & 4 & 0.25 \\
添油加醋 & oil vinegar & 66 & 23 & 0.522 \\
守株待兔 & rabbit & 64 & 29 & 0.31 \\
板上钉钉 & board & 64 & 37 & 0.162 \\
顺手牵羊 & sheep goat & 60 & 37 & 0.054 \\
呆若木鸡 & wood wooden chicken & 56 & 14 & 0.214 \\
生龙活虎 & dragon tiger & 54 & 40 & 0.375 \\
罄竹难书 & bamboo & 53 & 8 & 0 \\
九牛一毛 & nine ox fur feather & 49 & 17 & 0 \\
闭门造车 & cart car & 45 & 9 & 0.222 \\
老态龙钟 & dragon bell clock & 43 & 6 & 0 \\
行将就木 & wood & 39 & 17 & 0.118 \\
鼠目寸光 & mouse mice rat & 33 & 17 & 0.294 \\
蜻蜓点水 & dragonfly water & 33 & 11 & 0.455 \\
九死一生 & nine & 32 & 18 & 0.111 \\
鱼龙混杂 & fish dragon & 29 & 5 & 0.2 \\
三六九等 & three six nine & 19 & 5 & 0.6 \\
沾花惹草 & flower grass & 7 & 22 & 0.364 \\
鸡飞狗跳 & chicken dog & 7 & 19 & 0.211 \\
\end{tabular}
\caption{Idioms and blacklists in CIBB with training and test set frequency of each idiom, and blacklist trigger rate of WMT17 translation system.}
\end{table*}
\end{SChinese}
\end{CJK}


\end{document}